\documentclass[10pt,twocolumn,letterpaper]{article}

\usepackage[pagenumbers]{wacv}      % To produce the REVIEW version for the algorithms track
\usepackage{graphicx}
\usepackage{amsmath}
\usepackage{amssymb}
\usepackage{booktabs}
\usepackage{autobreak}
\usepackage{multirow}

\usepackage[pagebackref,breaklinks,colorlinks]{hyperref}

\usepackage[capitalize]{cleveref}
\crefname{section}{Sec.}{Secs.}
\Crefname{section}{Section}{Sections}
\Crefname{table}{Table}{Tables}
\crefname{table}{Tab.}{Tabs.}

\def\wacvPaperID{302} % *** Enter the WACV Paper ID here
\def\confName{WACV}
\def\confYear{2024}

\begin{document}

\title{3D-Aware Talking-Head Video Motion Transfer}

% \and

\author{
Haomiao Ni$^1$\quad\quad Jiachen Liu$^1$\quad\quad Yuan Xue$^2$ \quad\quad Sharon X. Huang$^1$\\
$^1$The Pennsylvania State University, University Park, PA, USA \\ 
$^2$The Ohio State University, Columbus, OH, USA \\
$^1${\tt\small\{hfn5052, jzl6493, suh972\}@psu.edu}\quad$^2${\tt\small{Yuan.Xue@osumc.edu}}
}

\maketitle

%%%%%%%%% ABSTRACT
\begin{abstract}
Motion transfer of talking-head videos involves generating a new video with the appearance of a subject video and the motion pattern of a driving video. Current methodologies primarily depend on a limited number of subject images and 2D representations, thereby neglecting to fully utilize the multi-view appearance features inherent in the subject video. In this paper, we propose a novel 3D-aware talking-head video motion transfer network, Head3D, which fully exploits the subject appearance information by generating a visually-interpretable 3D canonical head from the 2D subject frames with a recurrent network. A key component of our approach is a self-supervised 3D head geometry learning module, designed to predict head poses and depth maps from 2D subject video frames. This module facilitates the estimation of a 3D head in canonical space, which can then be transformed to align with driving video frames. Additionally, we employ an attention-based fusion network to combine the background and other details from subject frames with the 3D subject head to produce the synthetic target video. Our extensive experiments on two public talking-head video datasets demonstrate that Head3D outperforms both 2D and 3D prior arts in the practical cross-identity setting, with evidence showing it can be readily adapted to the pose-controllable novel view synthesis task.
\end{abstract}
\vspace{-2mm}

%%%%%%%%% BODY TEXT
\section{Introduction}
The task of transferring motion between talking-head videos, while maintaining the identity of the target subject, is a compelling research area with broad applications in special effects, entertainment, and video editing. Despite the significant progress in guided image-to-image synthesis, such as person image generation \cite{balakrishnan2018synthesizing,ma2017pose,pumarola2018unsupervised} and facial expression generation \cite{chen2020puppeteergan, kim2019u, ren2021pirenderer}, the challenge of capturing the temporal dynamics of motion in video-to-video transfer remains unsolved~\cite{chu2020learning,ni2023conditional}. Most current methods for talking-head video motion transfer use one subject image \cite{siarohin2019animating, siarohin2020first, siarohin2021motion} or a simple combination of a few subject images \cite{ha2020marionette,Ni_2023_WACV,wang2019few,wang2018video,wiles2018x2face} with 2D representations. These approaches may struggle to fully leverage the multi-view appearance information inherent in the subject video.

\begin{figure}[t]
    \centering
    \includegraphics[width=\linewidth]{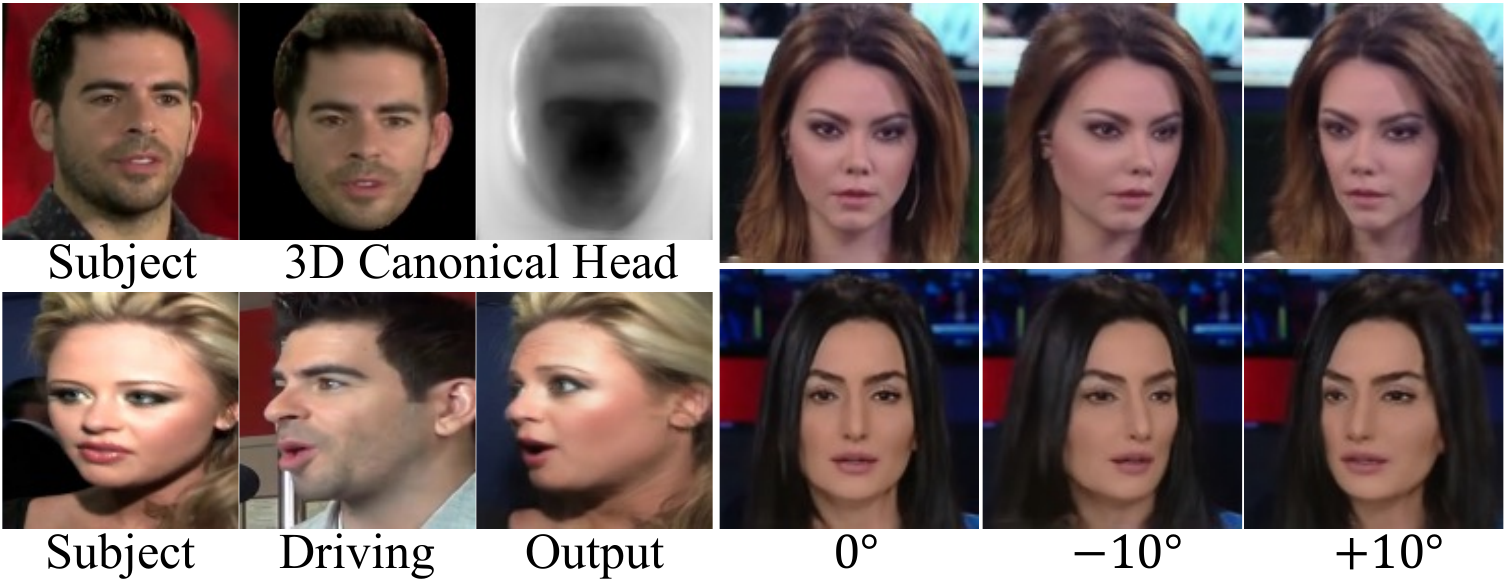}
    \caption{Illustration of advantages of our proposed Head3D. This 3D-aware framework can directly generate a 3D canonical head including RGB and depth map (top left), better deal with extreme poses (bottom left), and achieve pose-controllable novel view synthesis (right panel; changing yaw angle for the examples shown).}
    \label{fig:advantages}
    \vspace{-4mm}
\end{figure}

In this paper, we introduce \textit{Head3D}, a novel 3D-aware framework for transferring motion between talking-head videos. This framework operates in a self-supervised, non-adversarial training manner, and is capable of recovering 3D structural information (\ie, head pose and depth) from each 2D video frame through self-supervised 3D head geometry learning, without the need for a 3D graphical model of the human head. By mapping each selected subject video frame to a 3D canonical space, Head3D further estimates a 3D subject canonical head using a recurrent network. To synthesize the final video frames, Head3D employs an attention-based fusion mechanism to combine appearance features from the 3D subject head with background and other details (\eg, facial expression, shoulder) from the subject. 
%Unlike previous 3D-based methods \cite{drobyshev2022megaportraits,gafni2021dynamic,hong2022headnerf,siarohin2023unsupervised,wang2021one} that operate on implicit neural radiance field or canonical feature space, Head3D offers superior interpretability by explicitly modeling the 3D canonical head from the 2D subject video. 
{Unlike previous 3D-based methods that operate on the canonical feature space \cite{drobyshev2022megaportraits,siarohin2023unsupervised,wang2021one}, Head3D offers visual interpretability by explicitly modeling the 3D canonical head. Compared with NeRF-based methods \cite{gafni2021dynamic,hong2022headnerf,mildenhall2020nerf}, Head3D shows better generalization ability without the need to retrain the model on unseen faces.}
Moreover, with the generated 3D subject head, Head3D can effectively handle large pose changes or extreme poses and achieve novel view synthesis with user-provided pose inputs, as demonstrated in Fig.~\ref{fig:advantages}.
\textcolor{black}{
Our contributions are summarized as follows:
\begin{itemize}
% shorten our method and say something about other methods
% the advantage of 3D over 2D
%\item We propose a novel 3D-aware generative network Head3D to achieve talking-head video motion transfer by explicitly estimating a 3D canonical head to fully leverage the multi-view appearance information available in the subject video. This generated 3D head also enables Head3D to be easily applied to pose-controllable novel view synthesis. 
    %\item We introduce Head3D, a 3D-aware generative network for talking-head video motion transfer, which explicitly estimates a 3D canonical head without the need for any 3D shape priors. Unlike conventional 2D-based methods, our model can be readily adapted for pose-controllable novel view synthesis.
    \item {We introduce Head3D, a 3D-aware generative network for talking-head video motion transfer, which explicitly estimates a 3D canonical head without the need for any 3D shape priors.}
% compare with other 3D  
% \item We propose a self-supervised 3D head geometry learning strategy for head pose and depth prediction and produce the 3D canonical head from the 2D subject video with a recurrent network. An attention-based fusion network is further introduced to combine this 3D subject head with other information to generate the final synthetic video.
    %\item We employ a self-supervised 3D head geometry learning module with a recurrent network to generate a 3D canonical head from the 2D subject video. Compared to previous 3D-based methods, our resulting 3D canonical head representation is highly visual-interpretable.
    \item {We propose a self-supervised 3D head geometry learning module with a recurrent network to generate a 3D visually-interpretable canonical head from the 2D subject video.}
    %, which is more visually-interpretable compared with previous 3D methods operating on implict neural radiance field or canonical feature space. 
    % An attention-based fusion network is further introduced to combine this 3D subject head with other information to generate the final synthetic video.
%  stronger: outperform others under cross-identity
    %\item Through extensive experiments on two public talking-head video datasets, we demonstrate that Head3D outperforms other 2D- and 3D-based methods in the practical cross-identity motion transfer setting.
    \item {Comprehensive experiments demonstrate that our proposed Head3D outperforms other 2D- and 3D-based methods in the practical cross-identity motion transfer setting. Our model can also be easily adapted to pose-controllable novel view synthesis.}
\end{itemize}
}

\begin{figure*}[t]
    \centering
    \includegraphics[width=\textwidth]{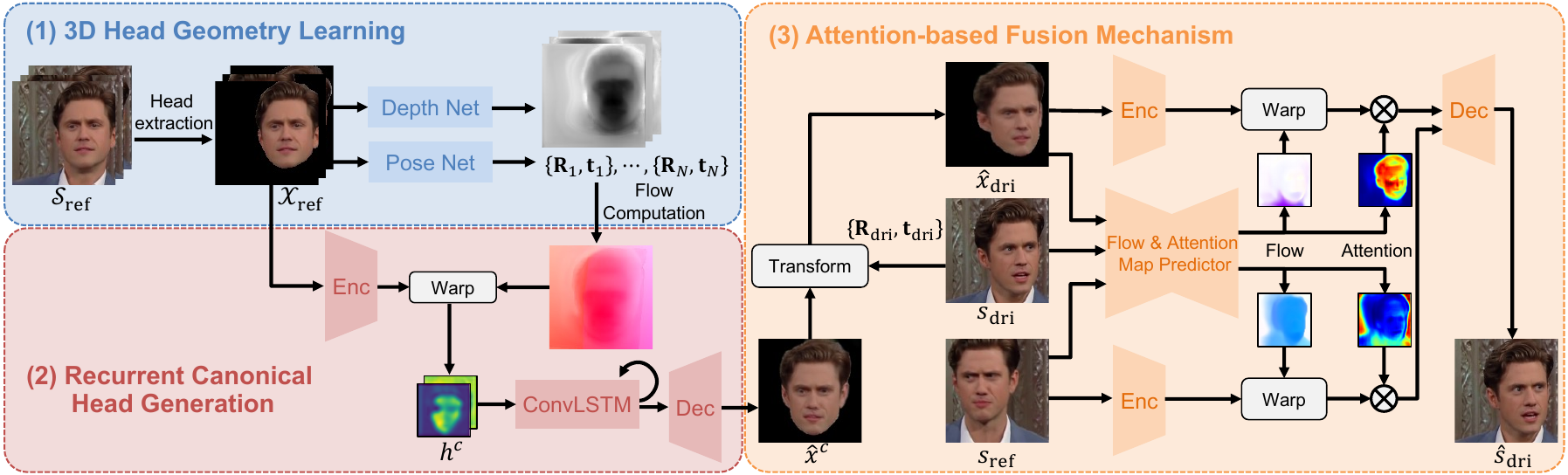}
    \caption{Illustration of the training framework of our proposed Head3D. During training, a set of reference subject frames $\mathcal{S}_\text{ref}$ and a driving frame $s_\text{dri}$ are randomly sampled from the same video $\mathcal{S}$. The final output frame $\hat{s}_\text{dri}$ will be used to compute a self-reconstruction loss against the driving frame $s_\text{dri}$ to train the network.
    {Note that these three stages are trained separately.}
    }
    \label{fig:training}
    \vspace{-2mm}
\end{figure*}

\section{Related Work}
According to whether 3D information is utilized during generation, talking-head video motion transfer methods can be categorized into 2D- or 3D-based frameworks. 

% 2D: one-shot and few-shot
\textbf{2D-based talking-head video motion transfer.} Based on whether to use multiple frames from the subject video, 2D-based methods can be further classified into one-shot \cite{siarohin2019animating, siarohin2020first, siarohin2021motion, tao2022structure,wang2022latent,zhao2022thin} and few-shot methods \cite{ha2020marionette,Ni_2023_WACV,wang2019few, wang2018video,wiles2018x2face,zakharov2019few,zakharov2020fast}. One-shot 2D methods, also known as image animation, focus on generating videos based on one given subject image and one driving video. Siarohin \etal \cite{siarohin2020first} proposed a general self-supervised first-order-motion framework (FOMM) to predict dense motion flow for animating arbitrary objects with learned keypoints and local affine transformations. In \cite{siarohin2021motion}, the authors further improved their network by modeling object movement through unsupervised region detection. Tao \etal \cite{tao2022structure} improved FOMM by introducing a deformable anchor model (DAM) to ensure that the object structure is well captured and preserved. 
However, these one-shot 2D methods are limited to using a single subject image, which makes it hard for them to utilize the multi-view appearance features of the subject when the subject video is available. 

Few-shot 2D methods instead utilize the subject video more effectively by synthesizing a video based on a number of subject video frames. Wang \etal \cite{wang2018video} proposed a video-to-video synthesis approach (vid2vid) under the generative adversarial learning framework \cite{goodfellow2014generative}, which produces one new video frame based on several previously generated images and the corresponding landmarks of the driving frame. In \cite{wang2019few}, they further proposed a few-shot vid2vid framework to learn how to synthesize videos of unseen subjects by leveraging a few example images of the target at test-time. 
Ha \etal \cite{ha2020marionette} proposed a few-shot face reenactment framework, MarioNETte, which employed image attention block, target feature alignment, and landmark transformer to address unseen identity and large-pose gaps. While few-shot methods have shown promising performance by utilizing appearance information from multiple frames, they operate only on 2D features and thus fail to fully exploit the multi-view information available in subject videos.

\textbf{3D-based talking-head video motion transfer.}
% 3D: nerf, 3DMM and others (DaGAN, face-vid2vid)
% different from previous implicit 3D head representation method, we explicit model the 3D canonical head and use it to help generation. More interpretability.
3D-based models have seen substantial progress in recent years. Some recent methods \cite{buhler2021varitex,geng20193d,ghosh2020gif,khakhulin2022realistic,liu20223d,taherkhani2023controllable} incorporate predefined shape models (\eg, 3DMM \cite{blanz1999morphable} or FLAME \cite{li2017learning}) to model 3D face for face manipulation. 
Liu \etal \cite{liu20223d} proposed 3D-FM GAN for 3D-controllable face manipulation by encoding both the input face image and a physically-based rendering of 3D edits into the latent space of StyleGAN \cite{karras2019style}. 
% However, these frameworks require fitting a 3D morphable model of the human face to every frame of a video and they are also limited in details of the shape. 
However, these methods depend on predefined 3D graphical models that may have limitations in modeling the unique shape details of different subjects. 
Other recent methods \cite{gafni2021dynamic,gao2022reconstructing,hong2022headnerf} instead used Neural Radiance Fields (NeRFs) \cite{mildenhall2020nerf} as a 3D representation of the human head. Gafni \etal \cite{gafni2021dynamic} proposed dynamic neural fields for modeling the appearance and dynamics of a human face tracked by 3DMM \cite{blanz1999morphable}. However, it can be hard for these NeRF-based models to generalize to unseen subject videos and they require fine-tuning or retraining when applied to new subjects. Some other methods \cite{drobyshev2022megaportraits,hong2022depth,tewari2022disentangled3d,wang2021one} are based on 3D geometrical transformation. Wang \etal \cite{wang2021one} proposed a one-shot free-view neural talking-head video synthesis model which represents a video using a sparse 3D keypoint representation. 
Hong \etal \cite{hong2022depth} introduced a self-supervised geometry learning method to automatically recover depth from face videos and leverage them to estimate sparse facial keypoints for talking head generation. Though also using 3D geometrical transformation, our proposed Head3D is different from these methods by explicitly modeling and visualizing the 3D canonical head estimated from the 2D subject video, thus providing an easily interpretable representation of the subject's head. 

\section{Methodology}
\label{sec:methods}
Figure \ref{fig:training} shows the training framework of our Head3D. 
%In general, Head3D is in an unsupervised training manner with self-reconstruction loss by utilizing several randomly sampled frames to restore another frame from the same video, without using any extra human annotations or involving adversarial training. 
In general, Head3D is trained in an unsupervised manner, using self-reconstruction loss to restore one video frame with several randomly sampled frames from the same video. This training process neither requires any human annotation nor involves adversarial training.
The training of Head3D consists of three stages: (1) 3D head geometry learning, (2) recurrent canonical head generation, and (3) attention-based fusion mechanism.  
{To ease the training, we train the modules in these three stages separately.}
Given a set of randomly sampled $N$ reference frames $\mathcal{S}_\text{ref}=\{s_1, s_2, \dots, s_N\}$ and a driving frame $s_\text{dri}$ from the same training video $\mathcal{S}$, in the first stage, we utilize a self-supervised 3D head geometry learning framework to train a depth network $F_\text{D}$ and a pose network $F_\text{P}$ for predicting the head pose and depth of each 2D video frame. 
During the second stage, we use a recurrent canonical head generation network that leverages ConvLSTM-based feature aggregation to create a 3D canonical head $\hat{x}^c$ incorporating warped reference frame features.
Finally, in stage three, we employ an attention-based fusion mechanism to synthesize each final output frame $\hat{s}_\text{dri}$ by combining head appearance from the canonical head $\hat{x}^c$, the background and other appearance details (\eg, neck and shoulder) from one randomly selected subject frame $s_\text{ref}$, and motion and expression information from the driving frame $s_\text{dri}$. Details of each component of our proposed framework are introduced as follows. 
%To ease the training, we train the modules in these three stages separately. 
More implementation details can be found in Sec~\ref{subsec:implement}. 

\subsection{3D Head Geometry Learning}
% BiSeNet
% face parsing network https://github.com/zllrunning/face-parsing.PyTorch
Given a talking-head video $\mathcal{S}$, we first randomly sample a set of $N$ reference frames $\mathcal{S}_\text{ref}=\{s_1, s_2, \dots, s_N\}$ and one driving frame $s_\text{dri}$ from $\mathcal{S}$. 
To recover the 3D geometry of the subject's head from a 2D talking-head video, we assume that videos are captured with a static perspective camera and that the subject's head can be treated as a rigid object. Our motivation is, by estimating a 3D head in canonical space, \ie, a 3D canonical head, the head region of each target video frame can be reconstructed by transforming the points of the 3D canonical head using a rigid pose transformation $\mathbf{P}=\{\mathbf{R}, \mathbf{t}\}\in\text{SE}(3)$. 
To only reconstruct the head part in subject video frames, we employ a pretrained face parsing network \cite{yu2021bisenet} to extract the facial and hair regions. This results in a set of reference head images $\mathcal{X}_\text{ref}=\{x_1, x_2, \dots, x_N\}$ and a driving head image $x_\text{dri}$. 

As shown in Fig~\ref{fig:training}, after head extraction, we apply a depth estimation network $F_\text{D}$ and a head pose prediction network $F_\text{P}$ to each frame in $\mathcal{X}_\text{ref}$ and $x_\text{dri}$ for estimating their depth maps $\mathcal{D}_\text{ref}=\{d_1, d_2, \dots, d_N\}$, $d_\text{dri}$, and their head poses $\mathcal{P}_\text{ref}=\{\mathbf{P}_1, \mathbf{P}_2, \dots, \mathbf{P}_N\}$, $\mathbf{P}_\text{dri}$. For each reference frame $x_i$ in $X_\text{ref}$, where $i=1, \dots, N$, we compute the corresponding canonical frame $x^c_i$ based on the image formation model in \cite{wu2020unsupervised}. Let pixel $q=(u, v, 1)$ be the homogeneous coordinate of one pixel in the reference frame $x_i$, and pixel $q^c=(u^c, v^c, 1)$ be the corresponding pixel in the canonical frame $x^c_i$. We can transform each pixel $q$ to $q^c$ to generate canonical frame $x^c$ by:
\begin{small}
\begin{equation}
\label{eq:warp_to_canon}
    q^c\propto K(\mathbf{R}^{T}(d[u, v]\cdot K^{-1}q-\mathbf{t}))\enspace,
\end{equation}
\end{small}
%the corresponding 3D point $Q=(Q_x, Q_y, Q_z)\in\mathbb{R}^3$ can be computed by:
% \begin{equation}
%     Q = m(u, v)\cdot K^{-1}q\enspace,
% \end{equation}
where $d[u, v]$ is the depth value of pixel $(u, v)$ in the depth map $d$, $\{\mathbf{R}, \mathbf{t}\}$ is the head pose of frame $x_i$, and $K$ is the camera intrinsic matrix, which can be computed by:
\begin{small}
\begin{equation}
\label{eq:K}
    K = \begin{pmatrix}
    f & 0 & c_u \\
    0 & f & c_v \\
    0 & 0 & 1
    \end{pmatrix}\enspace,
    \quad
    \left\{
    \begin{aligned}
    &c_u = \frac{W-1}{2}\\
    &c_v = \frac{H-1}{2}\\
    &f = \frac{W-1}{2\tan\frac{\theta_\text{FOV}}{2}}
    \end{aligned}
    \right.\enspace,
\end{equation}
\end{small}
where $H$ and $W$ are the height and width of image, $\theta_\text{FOV}$ is the field of view of the perspective camera. Following \cite{wu2020unsupervised}, we assume $\theta_\text{FOV}\approx10^{\circ}$ and the nominal distance of the subject from the camera is about 1m.
To simplify the training, we take the average of all the warped canonical frames $x^c_i$ to produce the final canonical frame $\Bar{x}^c$. We also apply $F_\text{D}$ to obtain its depth map $\Bar{d}^c$.
Similar to Eq.~\eqref{eq:warp_to_canon}, we can transform each pixel $q^c$ in the canonical frame to the pixel $q$ in the target frame by:
\begin{small}
\begin{equation}
\label{eq:warp_to_ref}
q\propto K(d^c[u^c, v^c]\cdot \mathbf{R}K^{-1}q^c+\mathbf{t})\enspace,
\end{equation}
\end{small}
where $q=(u, v, 1)$ is the homogeneous coordinate of one pixel in the target frame.
By applying $F_\text{P}$ to driving head frame $x_\text{dri}$ to estimate the head pose $\mathbf{P}_\text{dri}=\{\mathbf{R}_\text{dri}, \mathbf{t}_\text{dri}\}$, using Eq.~\eqref{eq:warp_to_ref}, we can transform the canonical frame $\Bar{x}^c$ to frame $\Bar{x}_\text{dri}$ with $\mathbf{P}_\text{dri}$.
Then we can train depth network $F_\text{D}$ and pose network $F_\text{P}$ with the following head reconstruction loss function:
\begin{small}
\begin{equation}
\label{eq:loss_geo}
    l_\text{geo} = ||\Bar{x}_\text{dri}-x_\text{dri}||_1 + \lambda_\text{sym}\mathcal{L}_\text{sym}(\Bar{x}^c) + \lambda_\text{D}\mathcal{L}_\text{D}(\Bar{d}^c)\enspace,
\end{equation}
\end{small}
where 
%$\mathcal{L}_\text{rec}$ is the loss measuring the difference between reconstructed frame $s_\text{out}$ and ground truth frame $s_\text{dri}$. Here we simply choose $L_1$ loss. 
%$\mathcal{L}_\text{sym}$ is designed to add symmetry constraint to the canonical head $\Bar{x}^c$. 
{$\mathcal{L}_\text{sym}$ is designed to ensure the estimated 3D head under the canonical pose by imposing symmetry constraint.}
Here $\mathcal{L}_\text{sym}=||\Bar{x}^c-\Bar{x}^{c^\prime}||_1$, where $\Bar{x}^{c^\prime}$ is the horizontally-flipped version of $\Bar{x}^c$. $\mathcal{L}_\text{D}$ is the depth smoothness loss used in \cite{godard2017unsupervised}. 
%Note that the depth map of canonical head $\Bar{m}^c$ can be easily obtained by applying depth estimation network $F_\text{D}$ to $\Bar{x}^c$. 
$\lambda_\text{sym}$ and $\lambda_\text{D}$ are balancing coefficients.
% $\Bar{x}^{c^\prime}$ is the horizontally-flipped version of $\Bar{x}^c$ and $\lambda$ is a balancing coefficient to add regularization loss for ensuring the canonical pose of $\Bar{x}^c$. Note that the depth map of $\Bar{x}^c$ can also be easily obtained by using depth estimation network $F_\text{D}$.

\begin{figure*}[t]
    \centering
    \includegraphics[width=0.8\linewidth]{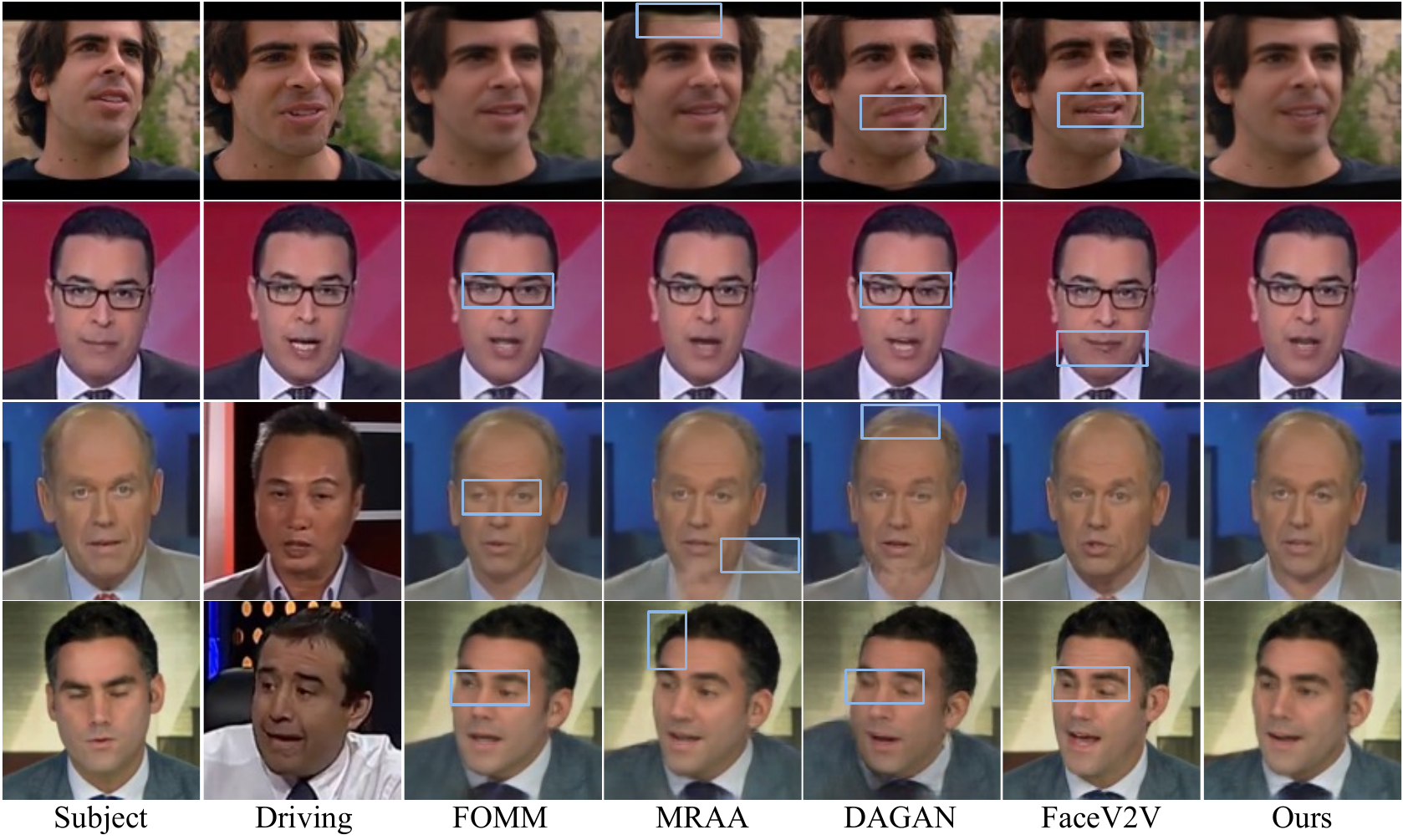}
    \caption{Qualitative comparison with state-of-the-art methods (FOMM \cite{siarohin2020first}, MRAA \cite{siarohin2021motion}, DAGAN \cite{hong2022depth} and FaceV2V \cite{wang2021one}). The top two rows are the results of self-reconstruction and the bottom two rows are that of cross-identity transfer. Artifacts and unnatural details are highlighted with blue boxes.}
    \label{fig:sota}
    \vspace{-3mm}
\end{figure*}

\subsection{Recurrent Canonical Head Generation}
% mean canonical head will only keep the intersection of every canonical head
The canonical head image $\Bar{x}^c$ is computed by averaging each transformed reference head frame in $\mathcal{X}_\text{ref}$. Thus $\Bar{x}^c$ is often blurry and not ready for the subsequent target frame generation. To produce a high-quality fine-grained canonical head, we propose a novel recurrent canonical head generation network to combine transformed reference frames. As shown in Fig.~\ref{fig:training}, for each reference head frame $x_i$, we utilize head image encoder $E_\text{H}$ to encode $x_i$ as feature $h_i$ and also use its corresponding depth $d_i$ and pose $\mathbf{P}_i$ to compute backward optical flow $f_{R_i\gets C}$ (\ie, warping from canonical head $x^c$ to reference head $x_i$) by:
%\vspace{-2mm}
\begin{small}
\begin{equation}
\label{eq:flow}
f_{R_i\gets C}[u, v] = (u, v)^T - (u^\text{c}, v^\text{c})^T\enspace,
\end{equation}
\end{small}
where $(u^\text{c}, v^\text{c})$ is one pixel in  $x^c$ and $(u, v)$ is the corresponding warped pixel in $x_i$ by applying Eq.~\eqref{eq:warp_to_ref} to each $(u^\text{c}, v^\text{c})$ with $\mathbf{P}_i$. Here we adopt the backward warping operation because it can be implemented efficiently in a differentiable manner using bilinear sampling \cite{jaderberg2015spatial}. 

We later apply flow $f_{R_i\gets C}$ to warp reference head feature $h_i$ to $h^c_i$. Then we employ a Convolutional LSTM \cite{shi2015convolutional} module $\Lambda$ to aggregate all $h^c_i$ to generate the final canonical head feature $h^c$. A head image decoder $\Omega_\text{H}$ is finally used to decode feature $h^c$ to be canonical head $\hat{x}^c$. Then we can apply $F_\text{D}$ to $\hat{x}^c$ and combine the estimated depth $\hat{d}^c$ with $\hat{x}^c$ to form a 3D canonical head. This 3D head helps to fully utilize the multi-view appearance information provided by different reference frames. By applying Eq.~\eqref{eq:warp_to_ref} to $\hat{x}^c$ using the driving head pose $\mathbf{P}_\text{dri}=\{\mathbf{R}_\text{dri}, \mathbf{t}_\text{dri}\}$, we transform $\hat{x}^c$ to the estimated driving head frame $\hat{x}_\text{dri}$. So we can train head image encoder $E_\text{H}$, image decoder $\Omega_\text{H}$, and ConvLSTM $\Lambda$ by the following head reconstruction loss function:
%\vspace{-2mm}
\begin{small}
\begin{equation}
    l_\text{head} = ||\hat{x}_\text{dri} - x_\text{dri}||_1\enspace.
\end{equation}
\end{small}
%\vspace{-2mm}
%where $x_\text{dri}$ is the ground truth driving head image segmented from $s_\text{dri}$.

\subsection{Attention-based Fusion Mechanism}
Because of modeling with rigid transformation, the estimated canonical head $\hat{x}^c$ can only describe movements of the whole head. To model facial expressions as well as the appearance and motion of non-head regions, we propose an attention-based fusion mechanism to combine $\hat{x}^c$, a randomly selected reference frame $s_\text{ref}$, and the motion and expression from driving frame $s_\text{dri}$ to produce the final target video frame $\hat{s}_\text{dri}$. As Fig.~\ref{fig:training} shows, we first transform the estimated canonical head $\hat{x}^c$ to driving head $\hat{x}_\text{dri}$ using the driving head pose $\mathbf{P}_\text{dri}=\{\mathbf{R}_\text{dri}, \mathbf{t}_\text{dri}\}$. 
%To provide information about background and other appearance details (\eg, shoulder and neck), we also randomly select a reference frame $s_\text{ref}$ as input. 
We then employ a frame encoder $E_\text{F}$ to represent $\hat{x}_\text{dri}$ and $s_\text{ref}$ as features $\hat{e}_\text{dri}$ and $e_\text{ref}$.
We also design a flow and attention map predictor $\Phi$, to which  $\hat{x}_\text{dri}$, $s_\text{ref}$ and $s_\text{dri}$ are fed, in order to estimate two backward warping feature flows $f_{\hat{x}_\text{dri}\gets s_\text{dri}}$ and $f_{s_\text{ref}\gets s_\text{dri}}$, and three attention maps $a_{\hat{x}_\text{dri}}$, $a_{s_\text{ref}}$ and $a_\text{dec}$. Then the combined feature ${e}_\text{out}$ can be computed by:
\begin{small}
\begin{equation}
\begin{split}
    {e}_\text{out} = & a_{\hat{x}_\text{dri}}\odot \mathcal{W}(\hat{e}_\text{dri}, f_{\hat{x}_\text{dri}\gets s_\text{dri}}) 
    + a_{s_\text{ref}}\odot \mathcal{W}(e_\text{ref}, f_{s_\text{ref}\gets s_\text{dri}}) 
    \\ & + a_\text{dec} \odot e_\text{dec}\enspace,
\end{split}
\end{equation}
\end{small}
where $\mathcal{W}(\cdot, \cdot)$ is backward warping, and $f_{\hat{x}_\text{dri}\gets s_\text{dri}}$ is used for warping estimated head $\hat{x}_\text{dri}$ to $s_\text{dri}$ for adding facial expression to $\hat{x}_\text{dri}$. $f_{s_\text{ref}\gets s_\text{dri}}$ is used for warping reference frame $s_\text{ref}$ to $s_\text{dri}$ for providing the background and other appearance details. $e_\text{dec}$ is the intermediate feature from decoder for synthesizing unseen novel regions. Attention maps $a_{\hat{x}_\text{dri}}$, $a_{s_\text{ref}}$, and $a_\text{dec}$ are designed 
to indicate which parts in the feature maps can be kept and which parts should be masked out. The sum of the attention weights for corresponding pixels in the three attention maps should be equal to 1. Finally we employ a frame decoder $\Omega_\text{F}$ to decode feature ${e}_\text{out}$ to target frame $\hat{s}_\text{dri}$. $\hat{s}_\text{dri}$ should be identical to $s_\text{dri}$ and thus we can train the frame encoder $E_\text{F}$, flow and attention map predictor $\Phi$, the frame decoder $\Omega_\text{F}$ using the following frame reconstruction loss:
\begin{small}
\begin{equation}
    l_\text{frame} = \mathcal{L}_\text{rec}(\hat{s}_\text{dri}, s_\text{dri})\enspace,
\end{equation}
\end{small}
where $\mathcal{L}_\text{rec}$ is the loss measuring the difference between reconstructed frame $\hat{s}_\text{dri}$ and ground truth frame $s_\text{dri}$. {Per \cite{siarohin2020first,siarohin2021motion}}, we implement $\mathcal{L}_\text{rec}$ using the perceptual loss \cite{johnson2016perceptual} based on pretrained VGG network \cite{simonyan2014very} {and also add the equivariance loss \cite{siarohin2020first} to stabilize the training}.

\subsection{Inference}
During testing, given one subject video $\mathcal{S}$ and one driving video $\mathcal{Y}=\{y_1, y_2, \dots, y_M\}$, we first randomly sample one reference image $s_\text{ref}$ and a set of reference frames $\mathcal{S}_\text{ref}$ from video $\mathcal{S}$, and estimate 3D canonical head $\hat{x}^c$ from $\mathcal{S}_\text{ref}$ through our proposed recurrent canonical head generation framework. Then for each driving frame $y_i$ in $\mathcal{Y}$, we adopt our attention-based fusion mechanism to combine $\hat{x}^c$, $s_\text{ref}$, and $y_i$ to generate the corresponding novel frame $\hat{s}_i$. The final target video $\hat{\mathcal{S}}=\{\hat{s}_1, \hat{s}_2, \dots, \hat{s}_M\}$ is generated in a frame-by-frame manner.

\textbf{Pose-controllable novel view synthesis.} %Our proposed Head3D can be easily extended to pose-controllable novel view synthesis task by manually inputting $\mathbf{P}_\text{dri}=\{\mathbf{R}_\text{dri}, \mathbf{t}_\text{dri}\}$ to transform the canonical head $\hat{x}^c$ to $\hat{x}_\text{dri}$ instead of obtaining $\mathbf{P}_\text{dri}$ from driving frame $s_\text{dri}$. 
Our proposed Head3D can be easily adapted to the pose-controllable novel view synthesis task by manually inputting the desired pose transformation, $\mathbf{P}_\text{dri}=\{\mathbf{R}_\text{dri}, \mathbf{t}_\text{dri}\}$, to generate the novel view $\hat{x}_\text{dri}$ from the canonical head representation, $\hat{x}^c$, rather than obtaining $\mathbf{P}_\text{dri}$ from the driving frame $s_\text{dri}$.
Then instead of inputting a $s_\text{dri}$ to the flow and attention map predictor $\Phi$, we input $\hat{x}_\text{dri}$ to the predictor, and the final output frame $\hat{s}$ will have the pose $\mathbf{P}_\text{dri}$.

\section{Experiments}

\begin{figure*}[t]
    \centering
    \includegraphics[width=0.88\textwidth]{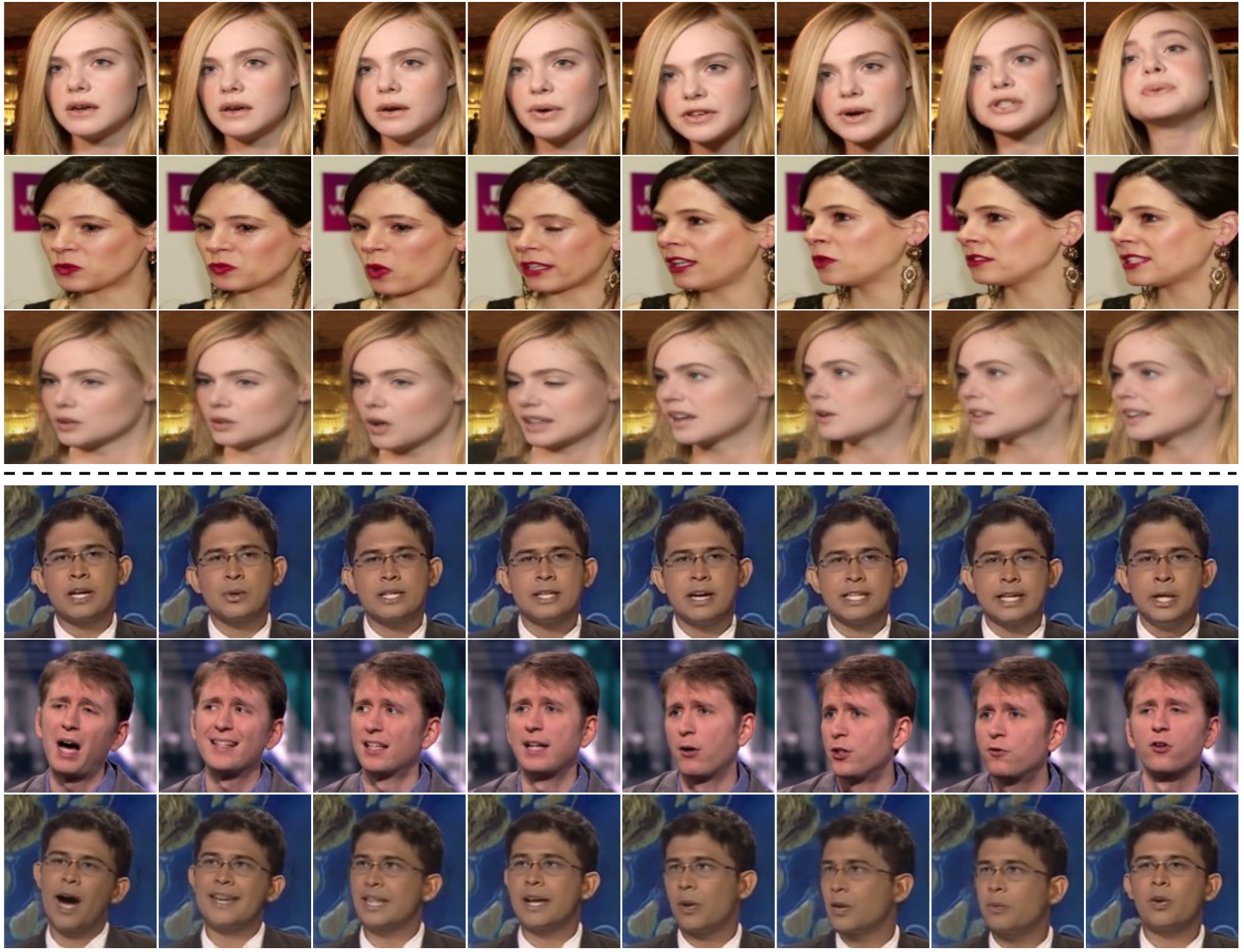}
    \caption{Examples of generated talking-head videos 
    %(top block) and dance videos (bottom block) 
    using our proposed Head3D. For each block, Head3D synthesizes the new video (3rd row) with the appearance from the subject video (1st row) and motions from the driving video (2nd row).}
    \label{fig:showcases}
    \vspace{-2mm}
\end{figure*}

\subsection{Datasets and Metrics}
\textbf{Datasets.}
We conduct comprehensive experiments on two public datasets: \textit{VoxCeleb} dataset \cite{nagrani2017voxceleb} and \textit{FaceForensics} dataset \cite{rossler2018faceforensics}. The VoxCeleb dataset includes 22,496 videos downloaded from Youtube. To simplify the training, we only keep 7,500 videos for training and 400 videos for testing. 
The FaceForensics dataset contains 1,004 videos of news briefings from different reporters. We find that models trained on the VoxCeleb dataset can be generalized well to this new dataset. So we only randomly choose 150 videos for testing without any additional training. Following the preprocessing approach in \cite{siarohin2020first}, we crop videos in these datasets to mainly keep the head regions and resize all video frames to $128\times 128$. Since the original videos in these datasets are long, we randomly select a short segment of 40 continuous frames from each video and use these selected short videos in our experiments.

\textbf{Metrics.} Following \cite{siarohin2020first}, we compute metrics based on two testing settings, \textit{self-reconstruction} and \textit{cross-identity transfer}. %For each setting, we randomly synthesize 100 videos. 
For self-reconstruction, we segment a video of the same subject into two non-overlapping clips. We use one clip as the subject video and the other as the driving video. In this setting, the driving video serves as ground truth. Similar to \cite{gafni2021dynamic,wang2022latent}, we compute the normalized mean $L_2$ distance and Learned Perceptual Image Patch Similarity (LPIPS) \cite{zhang2018unreasonable} metrics between self-reconstructed results and driving videos.
%we compute (1) the normalized mean $L_2$ distance to measure the frame-level difference between ground truth and generated videos and (2) average Euclidean distance (AED) \cite{siarohin2019animating} to evaluate the feature-level difference between the feature representation of the ground truth and generated frames. The facial feaure embedding is extracted by a trained face recognition network \cite{amos2016openface}. 
For cross-identity transfer, which is more practical in real-world applications, subject video and driving video are of different subjects in this setting. As there is no ground truth available, we conduct a paired user study to compare our model with state-of-the-art methods. {Specifically, we generated 100 videos for each baseline method on each dataset and paired them with videos produced by our model. We then invited 10 human evaluators to make judgments regarding the better video in each pair, considering aspects such as visual realism, motion accuracy, and identity consistency.}

\subsection{Implementation}
\label{subsec:implement}
\textbf{Model Implementation.} We employ a public pretrained face parsing network\footnote{\url{https://github.com/zllrunning/face-parsing.PyTorch}} to extract the head regions (face and hair) from each video frame. For 3D head geometry learning, we implement the depth network $F_\text{D}$ with a similar architecture used in \cite{khan2021efficient}. To stabilize the training, we add instance normalization layer \cite{ulyanov2016instance} to the decoder of $F_\text{D}$. 
We adopt a similar architecture to HopeNet \cite{Ruiz_2018_CVPR_Workshops} for the pose network $F_\text{P}$ in our implementation. The original HopeNet only predicts the yaw, pitch, and roll of the head (\ie, $\mathbf{R}$). To enable estimation of the 3D head translation $\mathbf{t}$, we modified the final layer of the network. To accelerate the training, we initialize most parameters in $F_\text{D}$ and $F_\text{P}$ with pretrained models provided in \cite{khan2021efficient} and \cite{Ruiz_2018_CVPR_Workshops}.
For the recurrent canonical head generation, we choose the architecture in \cite{johnson2016perceptual} to implement the head image encoder $E_\text{H}$ and decoder $\Omega_\text{H}$ with 2 downsampling blocks. We employ a one-layer ConvLSTM \cite{shi2015convolutional} to implement $\Lambda$. In our attention-based fusion mechanism, we also construct the frame encoder $E_\text{F}$ and decoder $\Omega_\text{F}$ using the same architecture as $E_\text{H}$ and $\Omega_\text{F}$. {The flow and attention map predictor $\Phi$ is built based on the flow predictor in MRAA \cite{siarohin2021motion}.} We slightly change its architecture to enable the prediction of three attention maps. 

\begin{figure*}[t]
    \centering
    \includegraphics[width=0.7\linewidth]{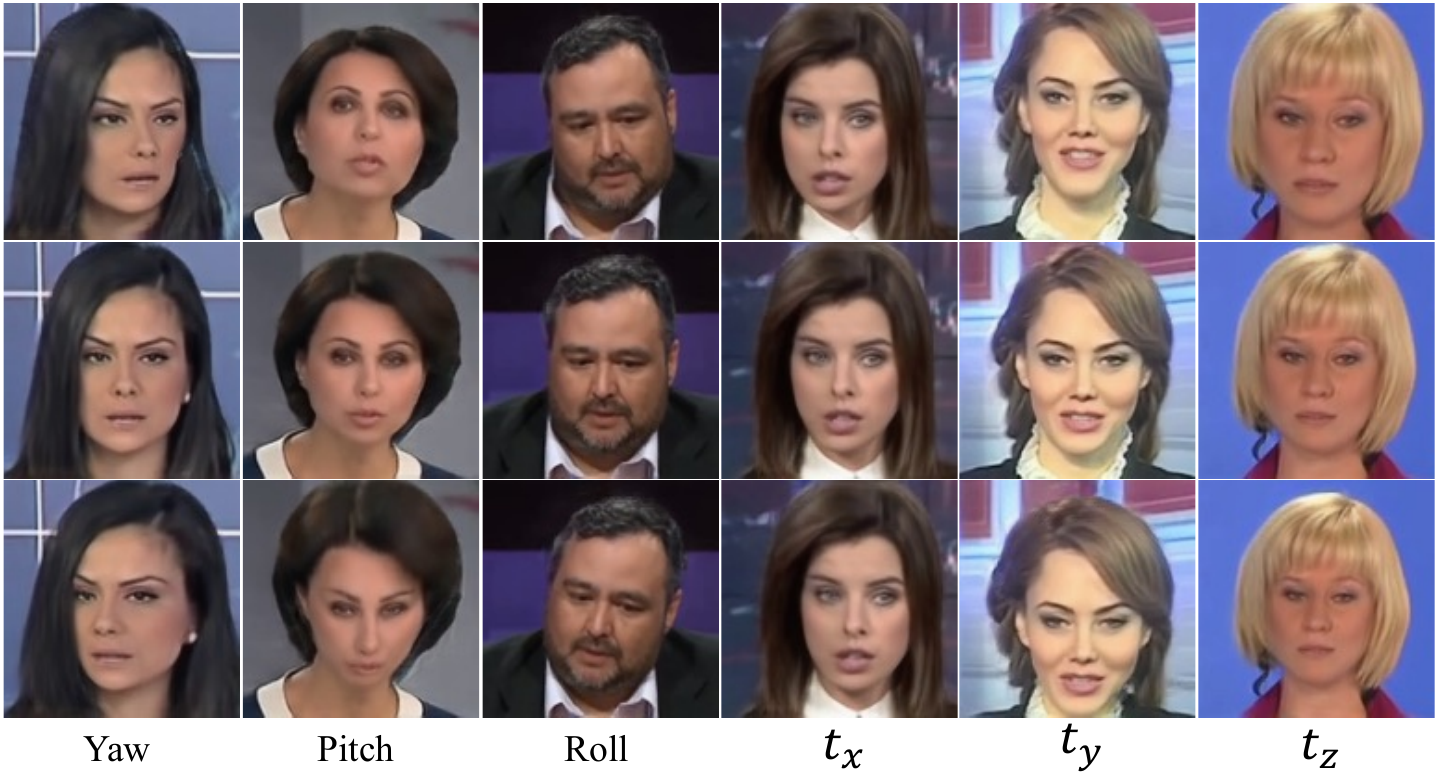}
    \caption{Examples of pose-controllable novel view synthesis. Each column demonstrates changing of a 3D rotation or translation parameter.}
    \label{fig:pose}
    \vspace{-3mm}
\end{figure*}

% TODO training details, loss, iteration, number of reference frames
As mentioned in Sec.~\ref{sec:methods}, the whole training process of Head3D includes three separate stages. In the first stage, we train the depth network $F_\text{D}$ and pose network $F_\text{P}$ through 3D head geometry learning. In the second stage, we train the head image encoder $E_\text{H}$, head decoder $\Omega_\text{H}$, and ConvLSTM $\Lambda$ for the recurrent canonical head generation. We finally train frame encoder $E_\text{F}$, frame decoder $\Omega_\text{F}$, and flow and attention map predictor $\Phi$ for the attention-based fusion mechanism in the third stage. 
We set batch size as 5 videos and use the Adam optimizer \cite{kingma2014adam} with $(\beta_1, \beta_2)=(0.5, 0.999)$ during all three-staged training. Unless otherwise specified, the number of reference frames is set to 5.
During 3D head geometry learning, we train $F_\text{D}$ and $F_\text{P}$ for 10 epochs. The learning rate of $F_\text{D}$ and $F_\text{P}$ is $2\times10^{-4}$ and $2\times10^{-5}$ and drops by 0.1 at epoch 5. The balancing parameter $\lambda_\text{sym}$ and $\lambda_\text{D}$ in Eq.~\ref{eq:loss_geo} are all set to be 0.1. When training recurrent canonical head generation, we train $E_\text{H}$, $\Omega_\text{H}$ and $\Lambda$ for 20 epochs with the learning rate of $2\times10^{-4}$ and drop learning rate by 0.1 at epoch 10. We train the attention-based fusion modules ($E_\text{F}$, $\Omega_\text{F}$ and $\Phi$) for 50 epochs with a fixed learning rate of $2\times10^{-4}$.

\textbf{Baseline Implementation.} We compare the proposed Head3D with three state-of-the-art motion transfer baseline models: 2D-based methods FOMM \cite{siarohin2020first} and MRAA \cite{siarohin2019animating}, and 3D-based methods DAGAN \cite{hong2022depth} and FaceV2V \cite{wang2021one}. We follow the default settings in the methods’
original implementations wherever possible\footnote{Due to the lack of official implementation, we implement FaceV2V with the code from \url{https://github.com/zhanglonghao1992/One-Shot_Free-View_Neural_Talking_Head_Synthesis}.} and retrain all the baselines using the same training videos on the VoxCeleb dataset as ours
with the same $128\times128$ resolution. 

% TODO FVD or other metrics
\begin{table}[t]
\centering
\caption{Comparison of proposed Head3D with state-of-the-art methods under the self-reconstruction setting on VoxCeleb and FaceForensics datasets.}
\vspace{1mm}
\resizebox{0.8\linewidth}{!}{%
\begin{tabular}{l|l|c|c}
\hline
Dataset                        & Method        & $L_2\downarrow$     & LPIPS$\downarrow$     \\ \hline
\multirow{4}{*}{VoxCeleb}      & FOMM \cite{siarohin2020first}          & 0.0114 & 0.0856 \\
                               & MRAA \cite{siarohin2021motion}          & \textbf{0.0108} & \textbf{0.0830} \\
                               & DAGAN \cite{hong2022depth}        & 0.0123 & 0.0885 \\ 
                               & FaceV2V \cite{wang2021one}
                               & 0.0186
                               & 0.0994 \\
                               \cline{2-4} 
                               & Head3D (Ours) & 0.0113 & 0.0855 \\ \hline
\multirow{4}{*}{FaceForensics} & FOMM \cite{siarohin2020first}          & 0.0102 & 0.0543 \\
                               & MRAA \cite{siarohin2021motion}         & \textbf{0.0075} & {0.0449} \\
                               & DAGAN \cite{hong2022depth}        & 0.0106 & 0.0490 \\ 
                               & FaceV2V \cite{wang2021one}
                               & 0.0119
                               & 0.0509 \\
                               \cline{2-4} 
                               & Head3D (Ours) & 0.0079 & \textbf{0.0442} \\ \hline
\end{tabular}%
}
\label{tab:sota_same}
\vspace{-1mm}
\end{table}

\begin{table}[t]
\centering
\caption{User preferences in the paired study: our approach \vs state-of-the-art methods under cross-identity setting on VoxCeleb and FaceForensics datasets.}
\vspace{1mm}
\resizebox{0.9\linewidth}{!}{%
\begin{tabular}{l|c|c}
\hline
Methods    & VoxCeleb (\%) & FaceForensics (\%) \\ \hline
Ours/FOMM \cite{siarohin2020first}  & \textbf{72}/28     & \textbf{68}/32          \\ 
Ours/MRAA \cite{siarohin2021motion}  & \textbf{57}/43     & \textbf{59}/41          \\
Ours/DAGAN \cite{hong2022depth} & \textbf{80}/20     & \textbf{86}/14          \\ 
Ours/FaceV2V \cite{wang2021one} & \textbf{53}/47 & \textbf{54}/46 \\
\hline
\end{tabular}%
}
\label{tab:sota_cross}
\end{table}

\subsection{Result Analysis}
\begin{figure*}[t]
    \centering
    \includegraphics[width=0.73\textwidth]{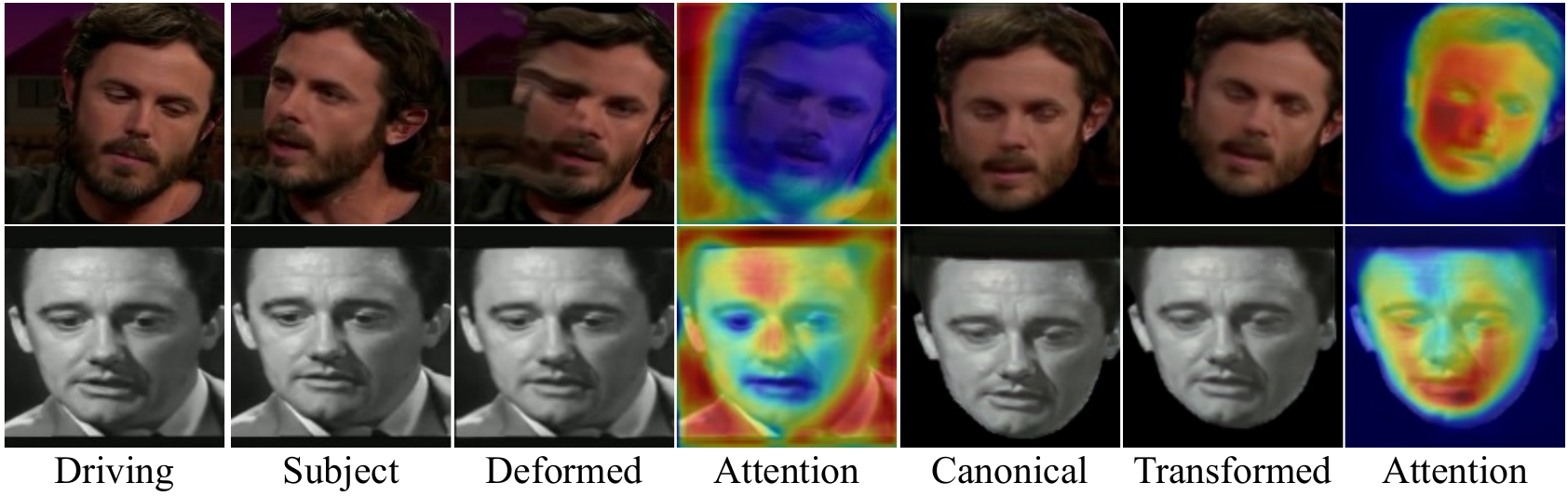}
    \caption{Illustration of the effect of attention maps in our proposed attention-based fusion mechanism. The red regions indicate a higher degree of attention, while the blue regions suggest a lower degree of attention. ``Deformed'' refers to applying warping flow $f_{s_\text{ref}\gets s_\text{dri}}$ to the subject frame and ``Transformed'' means applying driving pose $\mathbf{P}_\text{dri}$ to the canonical head.}
    \label{fig:attention}
    \vspace{-2mm}
\end{figure*}

% TODO Emphasize the advantage of 3D over 2D (do MRAA can't do), and self-reconstruction is not important in practice. Compare our 3D with other 3D methods (DAGAN, FaceV2V)
\textbf{Comparison with state-of-the-art methods.}
\textcolor{black}{
We compare our Head3D with state-of-the-art~(SOTA) methods under the self-reconstruction setting in Table~\ref{tab:sota_same}. As shown in Table~\ref{tab:sota_same}, Head3D achieves comparable or better performance when compared with the SOTA methods. While MRAA \cite{siarohin2021motion} performs better in most metrics under the self-reconstruction setting, our proposed Head3D outperforms it under the more practical cross-identity setting as shown in Table~\ref{tab:sota_cross}. Under the self-reconstruction setting, we speculate that the advantage of using the 3D canonical head in Head3D may not be apparent, as the head motion and pose changes are limited due to the subject and driving videos being clipped from the same original video.
When applied to the cross-identity motion transfer task, which typically involves larger head movements, Head3D benefits from leveraging the multi-view appearance information from the 3D canonical head, as is also shown in Fig.~\ref{fig:sota} and Fig.~\ref{fig:showcases}. More importantly, different from 2D-based FOMM and MRAA, Head3D can be easily applied to pose-controllable novel view synthesis, as shown in Fig.~\ref{fig:advantages} and Fig.~\ref{fig:pose}.
Additionally, unlike 3D-based DAGAN and FaceV2V, the canonical head representation in Head3D is visually interpretable, as shown in Fig.~\ref{fig:advantages} and Fig.~\ref{fig:attention}.
}

\textbf{Ablation Study.} 
% the effect of the number of reference frames
To analyze the effectiveness of each module in Head3D, we conduct an ablation study on the VoxCeleb dataset. Table~\ref{tab:aba} shows quantitative comparison results of the ablation study under the self-reconstruction setting. We first evaluate the effect of using different numbers of reference frames $N$. {Since ConvLSTM can utilize different number of reference frames during training and testing, in our experiments, we specifically train a model with 5 reference frames and then evaluate its performance with different number of reference frames during testing.}
Compared with our final model with 5 reference frames, using fewer frames ($N=1,3$) generated worse results while increasing the number of frames ($N=10$) can lead to better LPIPS but also longer inference time. So we choose $N=5$ as our default setting. To evaluate the effectiveness of proposed recurrent canonical head generation, 
we compare our model with [Head3D w/ $\bar{x}^c$], which employs the mean canonical head $\bar{x}^c$ instead of {the $\hat{x}^c$ generated by the recurrent network} to synthesize the final frames. 
One can observe that using mean canonical head $\bar{x}^c$ noticeably diminishes performance. The reason may be that $\bar{x}^c$ is generated by simply taking the average of all the canonical head images warped from reference frames, which makes it blurry and not capturing some important details. We also experiment with removing the canonical head input $\hat{x}_\text{dri}$ during attention-based fusion and evaluate this variant model [Head3D w/o $\hat{x}_\text{dri}$]. Without using the appearance information from $\hat{x}_\text{dri}$, the performance of [Head3D w/o $\hat{x}_\text{dri}$] decreases as Table~\ref{tab:aba} shows. 
\begin{table}[t]
\centering
\caption{Ablation Study under the self-reconstruction setting on VoxCeleb dataset.}
\vspace{1mm}
\resizebox{0.65\linewidth}{!}{%
\begin{tabular}{l|c|c}
\hline
Methods                         & \multicolumn{1}{c|}{$L_2\downarrow$}     & \multicolumn{1}{c}{LPIPS$\downarrow$}  \\ \hline
Head3D ($N=1$)                    & \multicolumn{1}{c|}{0.0117} & \multicolumn{1}{c}{0.0873} \\
Head3D ($N=3$)                    & \multicolumn{1}{c|}{0.0115} & \multicolumn{1}{c}{0.0880} \\
Head3D ($N=10$)                   & \multicolumn{1}{c|}{0.0116} & \multicolumn{1}{c}{\textbf{0.0839}} \\ \hline
Head3D w/ $\bar{x}^c$            & 0.0117                      & 0.0897                     \\
Head3D w/o $\hat{x}_\text{dri}$ & 0.0117                      & 0.0872                     \\ \hline
Head3D ($N=5$)                         & \textbf{0.0113}                      & {0.0855}                     \\ \hline
\end{tabular}%
}
\label{tab:aba}
\end{table}
% the effect of attention maps

We also illustrate the effectiveness of our proposed attention-based fusion mechanism by visualizing some examples of attention maps in Fig.~\ref{fig:attention}. As shown in Fig.~\ref{fig:attention}, when a significant pose difference exists between the subject and driving frames, as in the first row, our model will assign higher attention values to the transformed canonical head to synthesize facial areas. In cases where the poses are more similar, such as in the second row, our model will combine information from both the subject frame and canonical head to generate the facial regions.

\section{Limitation and Discussion}
\begin{figure}[t]
    \centering
    \includegraphics[width=0.75\linewidth]{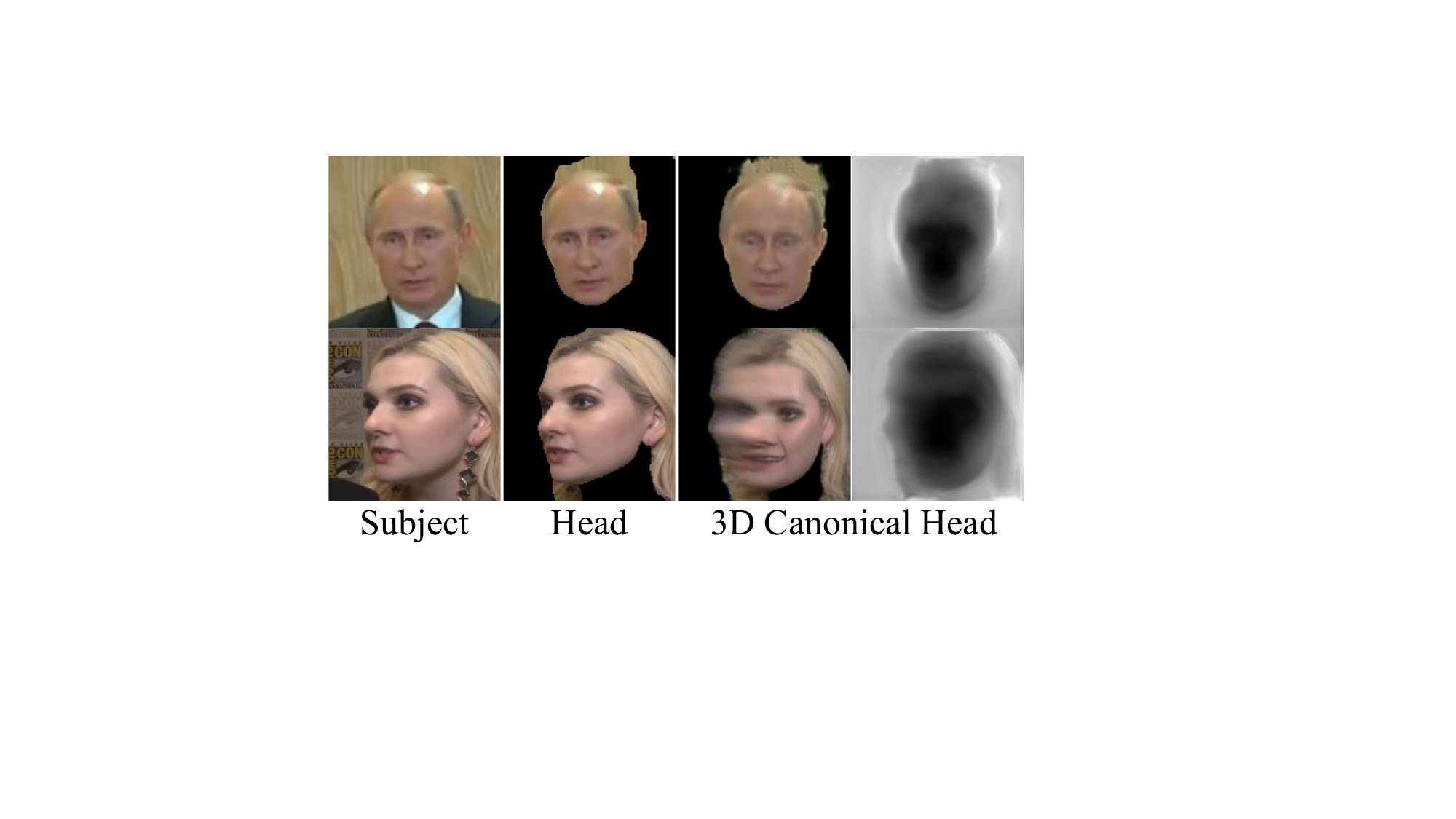}
    \caption{Examples of failure cases in canonical head estimation.}
    \label{fig:limitations}
    \vspace{-2mm}
\end{figure}

Head3D can achieve promising performance in most cases (see Fig.~\ref{fig:showcases} and Supp. videos). However, it still suffers from several limitations. First, our current framework employs an off-the-shelf face parsing network to segment the head regions from video frames. Imprecise segmentation performed by the pretrained network may result in inconsistent or incorrect extraction of head regions, which could further adversely impact the estimation of the 3D canonical head (see the 1st row in Fig.~\ref{fig:limitations}). Second, when the subject video only provides a single side-view of the person, it can be challenging to generate a high-quality canonical head (see the 2nd row in Fig.~\ref{fig:limitations}). Currently, our proposed attention-based fusion mechanism can mitigate these limitations by assigning lower attention values to incorrect details of the canonical head, thereby reducing their influence on the final synthesized output. In future work, we will investigate the use of a more robust pretrained face parsing network or incorporate an unsupervised face parsing model into the current framework to enable end-to-end training. {Recently, there has been a growing interest in high-resolution video generation \cite{drobyshev2022megaportraits}. We have provided a Supp. video at the $256\times256$ resolution, produced by our Head3D trained with different size parameters. In our subsequent research, we will also explore the video generation at the megapixel resolution such as $512\times512$.
}
%We also plan to enhance our recurrent canonical head generation framework to handle single side-view subject videos by leveraging some prior knowledge about human faces.
% rely on a pretrained face parsing network
% canonical head generation would be challengling if the subject video only contains the side-view of the subject.

% \textbf{Positive Negative Societal Impact.} Video motion transfer technologies could be misused for unethical purposes, \eg, creating fake news videos of celebrities \cite{yu2022generating}. Therefore, we will limit the usage of our method to research purposes only. Additionally, we intend to explore fake video detection techniques \cite{rahman2022qualitative} that can effectively identify fake videos generated using our proposed method.

\section{Conclusion}
In this paper, we present \textit{Head3D}, a novel 3D-aware approach for transferring motion in talking-head videos. Head3D capitalizes on the multi-view appearance information inherent in a 2D subject video by estimating a 3D canonical head using a recurrent network. We introduce a self-supervised 3D geometry learning module to predict pose and depth map, and an attention-based fusion network to generate the final synthesized video. The explicit modeling of a 3D canonical head in Head3D allows for easy application to novel view synthesis tasks using user-provided pose inputs. Comprehensive experiments on two public talking-head datasets demonstrate the state-of-the-art video motion transfer capabilities of Head3D. 

% In this paper, we propose a novel 3D-aware talking-head video motion transfer method, \textit{Head3D}, to fully utilize the multi-view appearance information from a 2D subject video by estimating a 3D canonical head with a recurrent network. To predict the 3D structure information, we propose a self-supervised 3D geometry learning module for pose and depth map estimation. An attention-based fusion network is further introduced to generate the final synthesized video. By explicitly modeling a 3D canonical head, our proposed Head3D can be easily applied to the novel view synthesis task with user-provided pose inputs. Comprehensive experiments on two public talking-head datasets demonstrate the state-of-the-art performance achieved by Head3D. 

{\small
\bibliographystyle{ieee_fullname}
\bibliography{egbib}

\begin{thebibliography}{10}\itemsep=-1pt

\bibitem{balakrishnan2018synthesizing}
Guha Balakrishnan, Amy Zhao, Adrian~V Dalca, Fredo Durand, and John Guttag.
\newblock Synthesizing images of humans in unseen poses.
\newblock In {\em Proceedings of the IEEE Conference on Computer Vision and
  Pattern Recognition}, pages 8340--8348, 2018.

\bibitem{blanz1999morphable}
Volker Blanz and Thomas Vetter.
\newblock A morphable model for the synthesis of 3d faces.
\newblock In {\em Proceedings of the 26th annual conference on Computer
  graphics and interactive techniques}, pages 187--194, 1999.

\bibitem{buhler2021varitex}
Marcel~C B{\"u}hler, Abhimitra Meka, Gengyan Li, Thabo Beeler, and Otmar
  Hilliges.
\newblock Varitex: Variational neural face textures.
\newblock In {\em Proceedings of the IEEE/CVF International Conference on
  Computer Vision}, pages 13890--13899, 2021.

\bibitem{chen2020puppeteergan}
Zhuo Chen, Chaoyue Wang, Bo Yuan, and Dacheng Tao.
\newblock Puppeteergan: Arbitrary portrait animation with semantic-aware
  appearance transformation.
\newblock In {\em Proceedings of the IEEE/CVF Conference on Computer Vision and
  Pattern Recognition}, pages 13518--13527, 2020.

\bibitem{chu2020learning}
Mengyu Chu, You Xie, Jonas Mayer, Laura Leal-Taix{\'e}, and Nils Thuerey.
\newblock Learning temporal coherence via self-supervision for gan-based video
  generation.
\newblock {\em ACM Transactions on Graphics (TOG)}, 39(4):75--1, 2020.

\bibitem{drobyshev2022megaportraits}
Nikita Drobyshev, Jenya Chelishev, Taras Khakhulin, Aleksei Ivakhnenko, Victor
  Lempitsky, and Egor Zakharov.
\newblock Megaportraits: One-shot megapixel neural head avatars.
\newblock In {\em Proceedings of the 30th ACM International Conference on
  Multimedia}, pages 2663--2671, 2022.

\bibitem{gafni2021dynamic}
Guy Gafni, Justus Thies, Michael Zollhofer, and Matthias Nie{\ss}ner.
\newblock Dynamic neural radiance fields for monocular 4d facial avatar
  reconstruction.
\newblock In {\em Proceedings of the IEEE/CVF Conference on Computer Vision and
  Pattern Recognition}, pages 8649--8658, 2021.

\bibitem{gao2022reconstructing}
Xuan Gao, Chenglai Zhong, Jun Xiang, Yang Hong, Yudong Guo, and Juyong Zhang.
\newblock Reconstructing personalized semantic facial nerf models from
  monocular video.
\newblock {\em ACM Transactions on Graphics (TOG)}, 41(6):1--12, 2022.

\bibitem{geng20193d}
Zhenglin Geng, Chen Cao, and Sergey Tulyakov.
\newblock 3d guided fine-grained face manipulation.
\newblock In {\em Proceedings of the IEEE/CVF conference on computer vision and
  pattern recognition}, pages 9821--9830, 2019.

\bibitem{ghosh2020gif}
Partha Ghosh, Pravir~Singh Gupta, Roy Uziel, Anurag Ranjan, Michael~J Black,
  and Timo Bolkart.
\newblock Gif: Generative interpretable faces.
\newblock In {\em 2020 International Conference on 3D Vision (3DV)}, pages
  868--878. IEEE, 2020.

\bibitem{godard2017unsupervised}
Cl{\'e}ment Godard, Oisin Mac~Aodha, and Gabriel~J Brostow.
\newblock Unsupervised monocular depth estimation with left-right consistency.
\newblock In {\em Proceedings of the IEEE conference on computer vision and
  pattern recognition}, pages 270--279, 2017.

\bibitem{goodfellow2014generative}
Ian Goodfellow, Jean Pouget-Abadie, Mehdi Mirza, Bing Xu, David Warde-Farley,
  Sherjil Ozair, Aaron Courville, and Yoshua Bengio.
\newblock Generative adversarial nets.
\newblock {\em Advances in neural information processing systems}, 27, 2014.

\bibitem{ha2020marionette}
Sungjoo Ha, Martin Kersner, Beomsu Kim, Seokjun Seo, and Dongyoung Kim.
\newblock Marionette: Few-shot face reenactment preserving identity of unseen
  targets.
\newblock In {\em Proceedings of the AAAI Conference on Artificial
  Intelligence}, volume~34, pages 10893--10900, 2020.

\bibitem{hong2022depth}
Fa-Ting Hong, Longhao Zhang, Li Shen, and Dan Xu.
\newblock Depth-aware generative adversarial network for talking head video
  generation.
\newblock In {\em Proceedings of the IEEE/CVF Conference on Computer Vision and
  Pattern Recognition}, pages 3397--3406, 2022.

\bibitem{hong2022headnerf}
Yang Hong, Bo Peng, Haiyao Xiao, Ligang Liu, and Juyong Zhang.
\newblock Headnerf: A real-time nerf-based parametric head model.
\newblock In {\em Proceedings of the IEEE/CVF Conference on Computer Vision and
  Pattern Recognition}, pages 20374--20384, 2022.

\bibitem{jaderberg2015spatial}
Max Jaderberg, Karen Simonyan, Andrew Zisserman, et~al.
\newblock Spatial transformer networks.
\newblock {\em Advances in neural information processing systems}, 28, 2015.

\bibitem{johnson2016perceptual}
Justin Johnson, Alexandre Alahi, and Li Fei-Fei.
\newblock Perceptual losses for real-time style transfer and super-resolution.
\newblock In {\em Computer Vision--ECCV 2016: 14th European Conference,
  Amsterdam, The Netherlands, October 11-14, 2016, Proceedings, Part II 14},
  pages 694--711. Springer, 2016.

\bibitem{karras2019style}
Tero Karras, Samuli Laine, and Timo Aila.
\newblock A style-based generator architecture for generative adversarial
  networks.
\newblock In {\em Proceedings of the IEEE/CVF conference on computer vision and
  pattern recognition}, pages 4401--4410, 2019.

\bibitem{khakhulin2022realistic}
Taras Khakhulin, Vanessa Sklyarova, Victor Lempitsky, and Egor Zakharov.
\newblock Realistic one-shot mesh-based head avatars.
\newblock In {\em European Conference on Computer Vision}, pages 345--362.
  Springer, 2022.

\bibitem{khan2021efficient}
Faisal Khan, Shahid Hussain, Shubhajit Basak, Joseph Lemley, and Peter
  Corcoran.
\newblock An efficient encoder--decoder model for portrait depth estimation
  from single images trained on pixel-accurate synthetic data.
\newblock {\em Neural Networks}, 142:479--491, 2021.

\bibitem{kim2019u}
Junho Kim, Minjae Kim, Hyeonwoo Kang, and Kwanghee Lee.
\newblock U-gat-it: unsupervised generative attentional networks with adaptive
  layer-instance normalization for image-to-image translation.
\newblock {\em arXiv preprint arXiv:1907.10830}, 2019.

\bibitem{kingma2014adam}
Diederik~P Kingma and Jimmy Ba.
\newblock Adam: A method for stochastic optimization.
\newblock {\em arXiv preprint arXiv:1412.6980}, 2014.

\bibitem{li2017learning}
Tianye Li, Timo Bolkart, Michael~J Black, Hao Li, and Javier Romero.
\newblock Learning a model of facial shape and expression from 4d scans.
\newblock {\em ACM Trans. Graph.}, 36(6):194--1, 2017.

\bibitem{liu20223d}
Yuchen Liu, Zhixin Shu, Yijun Li, Zhe Lin, Richard Zhang, and SY Kung.
\newblock 3d-fm gan: Towards 3d-controllable face manipulation.
\newblock In {\em Computer Vision--ECCV 2022: 17th European Conference, Tel
  Aviv, Israel, October 23--27, 2022, Proceedings, Part XV}, pages 107--125.
  Springer, 2022.

\bibitem{ma2017pose}
Liqian Ma, Xu Jia, Qianru Sun, Bernt Schiele, Tinne Tuytelaars, and Luc
  Van~Gool.
\newblock Pose guided person image generation.
\newblock {\em arXiv preprint arXiv:1705.09368}, 2017.

\bibitem{mildenhall2020nerf}
Ben Mildenhall, Pratul~P Srinivasan, Matthew Tancik, Jonathan~T Barron, Ravi
  Ramamoorthi, and Ren Ng.
\newblock Nerf: Representing scenes as neural radiance fields for view
  synthesis.
\newblock In {\em European Conference on Computer Vision}, pages 405--421.
  Springer, 2020.

\bibitem{nagrani2017voxceleb}
Arsha Nagrani, Joon~Son Chung, and Andrew Zisserman.
\newblock Voxceleb: a large-scale speaker identification dataset.
\newblock {\em arXiv preprint arXiv:1706.08612}, 2017.

\bibitem{Ni_2023_WACV}
Haomiao Ni, Yihao Liu, Sharon~X. Huang, and Yuan Xue.
\newblock Cross-identity video motion retargeting with joint transformation and
  synthesis.
\newblock In {\em Proceedings of the IEEE/CVF Winter Conference on Applications
  of Computer Vision (WACV)}, pages 412--422, January 2023.

\bibitem{ni2023conditional}
Haomiao Ni, Changhao Shi, Kai Li, Sharon~X Huang, and Martin~Renqiang Min.
\newblock Conditional image-to-video generation with latent flow diffusion
  models.
\newblock In {\em Proceedings of the IEEE/CVF Conference on Computer Vision and
  Pattern Recognition}, pages 18444--18455, 2023.

\bibitem{pumarola2018unsupervised}
Albert Pumarola, Antonio Agudo, Alberto Sanfeliu, and Francesc Moreno-Noguer.
\newblock Unsupervised person image synthesis in arbitrary poses.
\newblock In {\em Proceedings of the IEEE Conference on Computer Vision and
  Pattern Recognition}, pages 8620--8628, 2018.

\bibitem{ren2021pirenderer}
Yurui Ren, Ge Li, Yuanqi Chen, Thomas~H Li, and Shan Liu.
\newblock Pirenderer: Controllable portrait image generation via semantic
  neural rendering.
\newblock In {\em Proceedings of the IEEE/CVF International Conference on
  Computer Vision}, pages 13759--13768, 2021.

\bibitem{rossler2018faceforensics}
Andreas R{\"o}ssler, Davide Cozzolino, Luisa Verdoliva, Christian Riess, Justus
  Thies, and Matthias Nie{\ss}ner.
\newblock Faceforensics: A large-scale video dataset for forgery detection in
  human faces.
\newblock {\em arXiv preprint arXiv:1803.09179}, 2018.

\bibitem{Ruiz_2018_CVPR_Workshops}
Nataniel Ruiz, Eunji Chong, and James~M. Rehg.
\newblock Fine-grained head pose estimation without keypoints.
\newblock In {\em The IEEE Conference on Computer Vision and Pattern
  Recognition (CVPR) Workshops}, June 2018.

\bibitem{shi2015convolutional}
Xingjian Shi, Zhourong Chen, Hao Wang, Dit-Yan Yeung, Wai-Kin Wong, and
  Wang-chun Woo.
\newblock Convolutional lstm network: A machine learning approach for
  precipitation nowcasting.
\newblock {\em Advances in neural information processing systems}, 28, 2015.

\bibitem{siarohin2019animating}
Aliaksandr Siarohin, St{\'e}phane Lathuili{\`e}re, Sergey Tulyakov, Elisa
  Ricci, and Nicu Sebe.
\newblock Animating arbitrary objects via deep motion transfer.
\newblock In {\em Proceedings of the IEEE/CVF Conference on Computer Vision and
  Pattern Recognition}, pages 2377--2386, 2019.

\bibitem{siarohin2020first}
Aliaksandr Siarohin, St{\'e}phane Lathuili{\`e}re, Sergey Tulyakov, Elisa
  Ricci, and Nicu Sebe.
\newblock First order motion model for image animation.
\newblock {\em arXiv preprint arXiv:2003.00196}, 2020.

\bibitem{siarohin2023unsupervised}
Aliaksandr Siarohin, Willi Menapace, Ivan Skorokhodov, Kyle Olszewski, Jian
  Ren, Hsin-Ying Lee, Menglei Chai, and Sergey Tulyakov.
\newblock Unsupervised volumetric animation.
\newblock In {\em Proceedings of the IEEE/CVF Conference on Computer Vision and
  Pattern Recognition}, pages 4658--4669, 2023.

\bibitem{siarohin2021motion}
Aliaksandr Siarohin, Oliver~J Woodford, Jian Ren, Menglei Chai, and Sergey
  Tulyakov.
\newblock Motion representations for articulated animation.
\newblock In {\em Proceedings of the IEEE/CVF Conference on Computer Vision and
  Pattern Recognition}, pages 13653--13662, 2021.

\bibitem{simonyan2014very}
Karen Simonyan and Andrew Zisserman.
\newblock Very deep convolutional networks for large-scale image recognition.
\newblock {\em arXiv preprint arXiv:1409.1556}, 2014.

\bibitem{taherkhani2023controllable}
Fariborz Taherkhani, Aashish Rai, Quankai Gao, Shaunak Srivastava, Xuanbai
  Chen, Fernando de~la Torre, Steven Song, Aayush Prakash, and Daeil Kim.
\newblock Controllable 3d generative adversarial face model via disentangling
  shape and appearance.
\newblock In {\em Proceedings of the IEEE/CVF Winter Conference on Applications
  of Computer Vision}, pages 826--836, 2023.

\bibitem{tao2022structure}
Jiale Tao, Biao Wang, Borun Xu, Tiezheng Ge, Yuning Jiang, Wen Li, and Lixin
  Duan.
\newblock Structure-aware motion transfer with deformable anchor model.
\newblock In {\em Proceedings of the IEEE/CVF Conference on Computer Vision and
  Pattern Recognition}, pages 3637--3646, 2022.

\bibitem{tewari2022disentangled3d}
Ayush Tewari, Xingang Pan, Ohad Fried, Maneesh Agrawala, Christian Theobalt,
  et~al.
\newblock Disentangled3d: Learning a 3d generative model with disentangled
  geometry and appearance from monocular images.
\newblock In {\em Proceedings of the IEEE/CVF Conference on Computer Vision and
  Pattern Recognition}, pages 1516--1525, 2022.

\bibitem{ulyanov2016instance}
Dmitry Ulyanov, Andrea Vedaldi, and Victor Lempitsky.
\newblock Instance normalization: The missing ingredient for fast stylization.
\newblock {\em arXiv preprint arXiv:1607.08022}, 2016.

\bibitem{wang2019few}
Ting-Chun Wang, Ming-Yu Liu, Andrew Tao, Guilin Liu, Jan Kautz, and Bryan
  Catanzaro.
\newblock Few-shot video-to-video synthesis.
\newblock {\em arXiv preprint arXiv:1910.12713}, 2019.

\bibitem{wang2018video}
Ting-Chun Wang, Ming-Yu Liu, Jun-Yan Zhu, Guilin Liu, Andrew Tao, Jan Kautz,
  and Bryan Catanzaro.
\newblock Video-to-video synthesis.
\newblock {\em arXiv preprint arXiv:1808.06601}, 2018.

\bibitem{wang2021one}
Ting-Chun Wang, Arun Mallya, and Ming-Yu Liu.
\newblock One-shot free-view neural talking-head synthesis for video
  conferencing.
\newblock In {\em Proceedings of the IEEE/CVF conference on computer vision and
  pattern recognition}, pages 10039--10049, 2021.

\bibitem{wang2022latent}
Yaohui Wang, Di Yang, Francois Bremond, and Antitza Dantcheva.
\newblock Latent image animator: Learning to animate images via latent space
  navigation.
\newblock {\em arXiv preprint arXiv:2203.09043}, 2022.

\bibitem{wiles2018x2face}
Olivia Wiles, A Koepke, and Andrew Zisserman.
\newblock X2face: A network for controlling face generation using images,
  audio, and pose codes.
\newblock In {\em Proceedings of the European conference on computer vision
  (ECCV)}, pages 670--686, 2018.

\bibitem{wu2020unsupervised}
Shangzhe Wu, Christian Rupprecht, and Andrea Vedaldi.
\newblock Unsupervised learning of probably symmetric deformable 3d objects
  from images in the wild.
\newblock In {\em Proceedings of the IEEE/CVF Conference on Computer Vision and
  Pattern Recognition}, pages 1--10, 2020.

\bibitem{yu2021bisenet}
Changqian Yu, Changxin Gao, Jingbo Wang, Gang Yu, Chunhua Shen, and Nong Sang.
\newblock Bisenet v2: Bilateral network with guided aggregation for real-time
  semantic segmentation.
\newblock {\em International Journal of Computer Vision}, 129:3051--3068, 2021.

\bibitem{zakharov2020fast}
Egor Zakharov, Aleksei Ivakhnenko, Aliaksandra Shysheya, and Victor Lempitsky.
\newblock Fast bi-layer neural synthesis of one-shot realistic head avatars.
\newblock In {\em Computer Vision--ECCV 2020: 16th European Conference,
  Glasgow, UK, August 23--28, 2020, Proceedings, Part XII 16}, pages 524--540.
  Springer, 2020.

\bibitem{zakharov2019few}
Egor Zakharov, Aliaksandra Shysheya, Egor Burkov, and Victor Lempitsky.
\newblock Few-shot adversarial learning of realistic neural talking head
  models.
\newblock In {\em Proceedings of the IEEE/CVF International Conference on
  Computer Vision}, pages 9459--9468, 2019.

\bibitem{zhang2018unreasonable}
Richard Zhang, Phillip Isola, Alexei~A Efros, Eli Shechtman, and Oliver Wang.
\newblock The unreasonable effectiveness of deep features as a perceptual
  metric.
\newblock In {\em Proceedings of the IEEE conference on computer vision and
  pattern recognition}, pages 586--595, 2018.

\bibitem{zhao2022thin}
Jian Zhao and Hui Zhang.
\newblock Thin-plate spline motion model for image animation.
\newblock In {\em Proceedings of the IEEE/CVF Conference on Computer Vision and
  Pattern Recognition}, pages 3657--3666, 2022.

\end{thebibliography}
}
\end{document}